\documentclass[10pt,twocolumn,letterpaper]{article}

\usepackage{cvpr}              
\usepackage[pagebackref,breaklinks,colorlinks]{hyperref}
\hypersetup{allcolors=cvprblue}
\definecolor{cvprblue}{rgb}{0.21,0.49,0.74}

\usepackage[utf8]{inputenc}
\usepackage[T1]{fontenc}

\usepackage[colorlinks,breaklinks]{hyperref}
\usepackage{url}
\hypersetup{
  allcolors=black,   
  pagebackref=false, 
  pdfborder={0 0 0}
}

\def\paperID{*****} 
\def\confName{CVPR}
\def\confYear{2025}

\title{A Dataset Generation Scheme Based on Video2EEG-SPGN-Diffusion for SEED-VD}

\author{
Yunfei Guo\\
Chengdu Techman Software Co., Ltd.\\
Chengdu, China\\
{\tt\small guoyunfei@asu.edu.pl}
\and
Tao Zhang\\
Chengdu Techman Software Co., Ltd.\\
Chengdu, China\\
{\tt\small zhangtao@tme.com.cn}
\and
Wu Huang\textsuperscript{*}\\
Sichuan University\\
Chengdu, China\\
{\tt\small huangwu@scu.edu.cn}
\and
Yao Song\textsuperscript{*}\\
Sichuan University\\
Chengdu, China\\
{\tt\small yao.song@scu.edu.cn}
}

\begin{document}
\maketitle

\begin{abstract}
This paper introduces an open-source framework, Video2EEG-SPGN-Diffusion, that leverages the SEED-VD dataset to generate a multimodal dataset of EEG signals conditioned on video stimuli. Additionally, we disclose an engineering pipeline for aligning video and EEG data pairs, facilitating the training of multimodal large models with EEG alignment capabilities. Personalized EEG signals are generated using a self-play graph network (SPGN) integrated with a diffusion model. As a major contribution, we release a new dataset comprising over 1000 samples of SEED-VD video stimuli paired with generated 62-channel EEG signals at 200 Hz and emotion labels, enabling video-EEG alignment and advancing multimodal research. This framework offers novel tools for emotion analysis, data augmentation, and brain-computer interface applications, with substantial research and engineering significance.
\end{abstract}

\section{Introduction}
\subsection{Background}

The scaling law indicates that model performance scales with data volume, model size, and computational resources \citep{kaplan2020scaling}. However, electrophysiological data like EEG suffers from scarcity due to high acquisition costs and complexity \citep{sato2024scaling}. For instance, a model trained on 175 hours of EEG data achieved 48\% top-1 accuracy in phrase classification, versus 2.5\% with 10 hours \citep{sato2024scaling}. As brain-computer interface (BCI) devices proliferate, EEG data volume is expected to grow, raising privacy concerns. EEG can encode sensitive cognitive and emotional information, risking identity inference \citep{sidebottom2022eeg}, emotional states \citep{zhao2021emotion}, or covert intentions \citep{makin2020machine}. Centralized repositories increase breach vulnerabilities \citep{faro2024data}, necessitating robust data governance and encryption \citep{iyad2023neuroethics, nuffel2024bci}.

\subsection{Research Motivation}

\textbf{Can generative data address EEG scarcity and privacy risks?} Generative models (e.g., GANs, VAEs, diffusion models) can mitigate data scarcity by producing synthetic EEG for training, improving classification tasks \citep{alhaddad2023generative, song2024generating}. They also reduce privacy risks by generating data without identifiable information, minimizing exposure to breaches or cyber attacks \citep{ddpm2023, 9206683, 10517483}. Privacy-preserving GANs have successfully classified EEG while protecting identities \citep{01169w}.

\subsection{Research Objectives and Contributions}

We introduce an open-source Video2EEG-SPGN-Diffusion framework using the SEED-VD dataset to generate personalized EEG signals under video conditions, addressing data scarcity and privacy. The framework simulates video-watching scenarios, producing a new dataset of approximately 1000+ samples (256x256 video segments at 1-2 Hz, 62-channel EEG at 200 Hz, emotion labels) with no real personal data, ensuring safe sharing. We also disclose an engineering pipeline for video-EEG alignment, the first to simulate human neurophysiological responses to audio-visual stimuli, supporting multimodal model training for BCI and cross-modal research.

\subsection{Contributions}
\begin{itemize}
\item Open-sourcing the Video2EEG-SPGN-Diffusion framework for video-based EEG generation.
\item Releasing a new dataset, aligning video stimuli and generated EEG.
\item Providing a reproducible video-EEG alignment pipeline for multimodal model development.
\end{itemize}

\section{Related Work}
\subsection{Reliability of Diffusion Models Compared to GANs and VAEs}
Diffusion models generate data through iterative denoising, offering superior stability and sample quality compared to GANs, which may suffer from mode collapse, and VAEs, which often produce blurred outputs \citep{ho2020denoising}. In electrophysiological data generation, diffusion models excel at capturing complex temporal dynamics, such as EEG frequency bands and phase-amplitude coupling \citep{song2024generating}. Their robustness makes them more reliable than GANs and VAEs for generating electrophysiological signals.

\subsection{Generating Synthetic Neurophysiological Data}
The study \citep{ddpm2023} demonstrates the efficacy of denoising diffusion probabilistic models (DDPM) in generating EEG, ECoG, MEG, and LFP signals, effectively capturing key statistical and dynamic features of real data. Additionally, the work presented in \citep{EEGCiD_EEG_Condensation_Into_Diffusion_Model} introduces EEGCiD, a novel EEG condensation framework that leverages a deterministic denoising diffusion implicit model (DDIM) to synthesize a highly information-concentrated dataset. EEGCiD employs a transformer architecture with a spatial and temporal self-attention block (STSA) to enhance EEG knowledge modeling, optimizing latent codes to match feature distributions between synthetic and original datasets, addressing storage and privacy concerns in EEG-based applications.

\subsection{SEED-DV Dataset}
The SEED-DV dataset includes 62-channel EEG signals (200 Hz sampling rate) recorded from 15 participants viewing 72 video segments, with emotion labels such as happy, sad, neutral, and fear. The video stimuli are 2-4 minute segments \citep{liu2024eegvideo}.

\subsection{Self-play Fusion Graph }
Building on recent advances in reliability-enhanced Brain–Computer Interfaces via mixture-of-graphs-driven information fusion \cite{DAI2025103069}, we integrate three key strategies into the Video2EEG-SPGN-Diffusion framework. First, the self-play fusion strategy is employed to dynamically adjust the weight of various graph-based features, enabling robust information fusion under varying signal conditions. Second, the Graph-DA (graph-based data augmentation) approach introduces perturbations like noise addition and channel dropout, enhancing the model’s generalization and stability by diversifying training conditions. Finally, filter-driven multi-graph construction utilizes filter banks across different EEG frequency bands to build complementary graphs, capturing spatial, spectral, and temporal dependencies. These combined mechanisms strengthen the robustness of the SPGN, which generates EEG signals conditioned on video stimuli, offering a powerful framework for multimodal data generation, emotion analysis, and brain-computer interface applications.

\section{Methods}

\subsection{Video2EEG-SPGN-Diffusion Framework}

The Video2EEG-SPGN-Diffusion framework utilizes the SEED-VD dataset, comprising 62-channel EEG signals sampled at 200 Hz, video stimuli at 1–2 Hz, and emotion labels (happy, sad, neutral, fear). This dataset enables the generation of personalized EEG signals conditioned on video inputs, supporting applications in data augmentation and brain-computer interfaces.

\begin{figure*}[t!]
\centering
\includegraphics[scale=0.39]{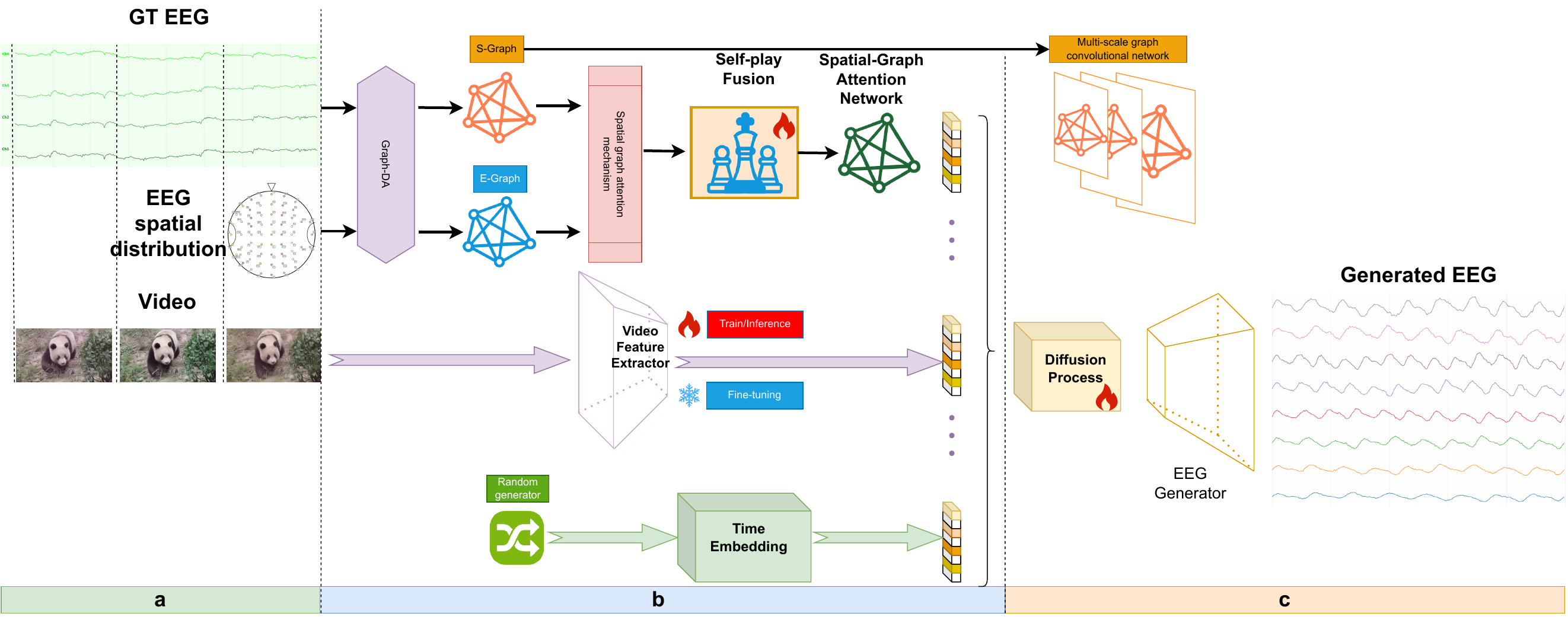}
\caption{Video2EEG-SPGN-Diffusion Framework for generating synthetic EEG signals from video stimuli, depicted in three stages: (a) Input Stage: Video inputs, ground-truth EEG (GT EEG) from SEED-VD, and EEG spatial distribution representing electrode scalp arrangement. (b) Feature Processing and Fusion: Video frames are processed by CLIP ViT-L/14, with features fused with subject-specific information via a self-play graph network (SPGN), incorporating graph-based data augmentation, electrode and signal graphs, and spatial-graph attention. (c) EEG Generation: Fused features are input into a Denoising Diffusion Probabilistic Model (DDPM) to generate 62-channel EEG signals at 200 Hz through iterative diffusion.}
\label{fig1}
\end{figure*}

The framework’s architecture, illustrated in Fig.~\ref{fig1}, processes video inputs, ground-truth EEG from the SEED-VD dataset, and EEG spatial distribution to generate synthetic EEG signals. The workflow consists of three main stages: input processing, feature fusion, and EEG generation.

In the input processing stage, video frames are extracted using the CLIP ViT-L/14 model, resized to 256×256 at 1–2 Hz, yielding feature vectors $v \in \mathbb{R}^{4 \times 768}$ over 2 seconds (0.5 seconds per frame). Subject-specific information is encoded via the CLIP text encoder as $e_{\text{text}} \in \mathbb{R}^{77 \times 768}$. Optionally, prior EEG data (62 channels, 200 Hz, 2 seconds) is processed using GLMNet to produce $e_{\text{eeg-prior}} \in \mathbb{R}^{512}$. These modalities are fused through cross-modal attention, formulated as $\text{Attn}(Q_v, K_{\text{text}}, V_{\text{text}})$, where $Q_v = W_Q \cdot v$, $K_{\text{text}} = W_K \cdot e_{\text{text}}$, and $V_{\text{text}} = W_V \cdot e_{\text{text}}$, resulting in fused features $e_{\text{fused}} \in \mathbb{R}^{4 \times 512}$.

The self-play graph network (SPGN) enhances feature representation through integrated components. Graph-based data augmentation (Graph-DA) introduces perturbations (noise addition, channel dropout, time offsets, amplitude scaling) at a 0.3 ratio to improve robustness. The electrode graph (E-Graph) is based on physical electrode distances, while the signal graph (S-Graph) leverages signal correlations from multi-band filter banks (delta, theta, alpha, beta, gamma). Multi-scale graph convolution, applied over 5 layers with a hidden dimension of 256, is augmented by a spatial-graph attention mechanism to capture spatial dependencies. Self-play fusion employs adversarial game optimization with a fusion weight of 0.001, ensuring optimal integration of graph-based features.

In the EEG generation stage, a Denoising Diffusion Probabilistic Model (DDPM) generates 62-channel EEG signals at 200 Hz, conditioned on SPGN outputs. The diffusion process uses 1000 steps during training and 50 steps during inference, following a cosine noise schedule with beta values from 1e-4 to 0.02, producing signals in a 62 × 200 format.

Training involves preprocessing EEG signals with bandpass filtering (0.5–40 Hz) and normalization to [-1,1], video normalization to 256×256, and text encoding via CLIP ViT-L/14. Optimization employs multiple loss functions: diffusion loss for denoising, adversarial loss for game optimization, frequency domain loss for spectral consistency, spatial loss for channel relationships, and temporal loss for sequence matching. Training parameters include 100 epochs, a batch size of 4, a learning rate of 1e-5 with the Adam optimizer, and gradient clipping to limit adversarial loss below 1000.

This open-source framework enables the generation of personalized video-conditioned EEG signals, facilitating advanced data augmentation and brain-computer interface applications.

\subsection{Video-EEG Data Alignment Engineering Implementation Pipeline}

We present a reproducible engineering pipeline for aligning video and EEG data pairs, designed to support the training of multimodal large models with robust EEG signal alignment capabilities. This pipeline encompasses data preprocessing, model training, signal generation, data alignment, and open-source implementation, ensuring seamless integration and scalability.

The pipeline commences with data preprocessing. Videos are read using OpenCV, converted from BGR to RGB, resized to $224 \times 224$, and normalized to the range $[0,1]$. Each video is stored as a tensor of shape $(B,T,C,H,W)$ with $250$ frames. EEG signals are processed into $14$ channels with $250$ sampling points at $250$ Hz, with values constrained to the range $[-10,10]$. The resulting data are stored in \texttt{NPZ} format, including both \texttt{video} and \texttt{eeg} arrays.

Model training is based on the Video2EEG-SGGN-Diffusion architecture, which integrates a Spatial-Graph Graph Neural Network (SGGN) with a denoising diffusion model. The training employs $1000$ diffusion steps, a hidden dimension of $256$, and a noise removal mechanism to refine EEG generation. Graph-based data augmentation (Graph-DA) is also applied to improve robustness and generalization. Both training and inference adopt the diffusion framework in a noise denoising regime.

Signal generation directly takes preprocessed video tensors as inputs, with the SGGN-Diffusion model producing aligned EEG outputs of size $14 \times 250$. Data alignment is maintained by synchronizing the $250$ video frames with their corresponding EEG sequences. The aligned samples are stored in \texttt{NPZ} format, each entry containing both video $(250 \times 3 \times 224 \times 224)$ and EEG $(14 \times 250)$ signals.

Quality evaluation includes multiple objective metrics: mean squared error (MSE), mean absolute error (MAE), temporal correlation, and frequency-domain similarity across EEG bands. These measures ensure fidelity of the generated EEG relative to target distributions.

The open-source implementation is built on PyTorch, integrating libraries such as OpenCV and NumPy, and provides reproducible scripts for preprocessing, model training, dataset construction, and evaluation. It is hosted on GitHub with documentation and example datasets to support downstream applications such as multimodal learning and emotion classification.

\subsection{Public Dataset Construction Detailed Scheme}

\subsubsection{General Scheme}
To address the challenges of data scarcity and privacy concerns in electrophysiological research, we construct a novel multimodal dataset by leveraging video stimuli from the SEED-VD dataset and synthetic EEG signals generated via the Video2EEG-SPGN-Diffusion framework. This dataset facilitates precise alignment between video content and corresponding EEG responses, enabling advanced multimodal investigations in areas such as emotion recognition, brain-computer interfaces, and cross-modal learning. By synthesizing EEG signals that mimic real neurophysiological patterns without incorporating sensitive personal data, the dataset ensures ethical compliance while supporting scalable model training.

\subsubsection{Detailed Scheme}
The video sources are derived from the SEED-VD dataset, which comprises EEG recordings from 15 participants exposed to 72 video segments, each lasting 2--4 minutes and designed to elicit specific emotions including happiness, sadness, neutrality, and fear. This results in approximately 1080 samples, providing a diverse foundation for multimodal data generation.

For EEG generation, we employ the Video2EEG-SPGN-Diffusion model to produce corresponding 62-channel EEG signals at a 200 Hz sampling rate, with each segment spanning 1 second (200 samples) to align with the model's optimized signal length. Personalization is achieved by conditioning the generation process on subject-specific information from the SEED-VD dataset, such as text embeddings describing demographic details (e.g., age, gender, and emotional arousal levels) and associated emotion labels. This conditioning ensures that the synthesized EEG signals are consistent with the induced emotional states, enhancing their biological plausibility and relevance to real-world scenarios.

The dataset construction integrates these elements into a cohesive structure, where each sample consists of video segments resized to \(256 \times 256\) resolution at a 1--2 Hz frame rate, paired with the generated EEG data (\(62 \times 200\) dimensions at 200 Hz), and annotated with emotion labels (happy, sad, neutral, or fear). Drawing from the SEED-VD foundation, the dataset encompasses over 1000 samples, offering substantial scale for research applications. Alignment between video frames and EEG signals is meticulously enforced using timestamps, achieving synchronization at intervals such as 0.5 seconds per time step to preserve temporal coherence. To validate the dataset's fidelity, we calculate power spectrum similarities across key frequency bands (e.g., \(\delta\), \(\theta\), \(\alpha\), \(\beta\), and \(\gamma\)) and assess emotion classification accuracy on the generated EEG, confirming consistency with the original SEED-VD dataset's statistical and functional properties.

\begin{table}[t!]
\centering
\caption{Metadata for Representative Generated EEG Dataset Samples}
\label{tab:dataset_metadata}
\begin{tabular}{l c c}
\toprule
\textbf{Metric} & \textbf{Sample 1} & \textbf{Sample 2} \\
\midrule
Subject ID & 1 & 2 \\
Video ID & 2 & 1 \\
Start Time (s) & 117.61 & 404.79 \\
Video Shape & $[60,224,224,3]$ & $[60,224,224,3]$ \\
EEG Shape & \(62 \times 200\) & \(62 \times 200\) \\
Duration (s) & 5.0 & 5.0 \\
Sampling Rate (Hz) & 200 & 200 \\
\bottomrule
\end{tabular}
\end{table}

To provide insight into the dataset's structure, Table~\ref{tab:dataset_metadata} presents metadata for two representative samples generated using the Video2EEG-SPGN-Diffusion framework. The table includes subject and video identifiers, sample index, start time, and original metadata detailing video and EEG dimensions, duration, and sampling rate. The samples correspond to subjects 1 and 2, paired with videos 2 and 1, respectively, with each video segment comprising 60 frames at \(224 \times 224 \times 3\) (RGB) and EEG signals at \(62 \times 200\) (62 channels, 200 samples at 200 Hz). The duration of 5 seconds for each sample indicates the temporal window used for generation, with start times (e.g., 117.61s and 404.79s) reflecting specific segments within the original SEED-VD videos. This metadata ensures traceability and reproducibility, allowing researchers to verify the alignment and characteristics of the generated data.

To evaluate the quality and performance of the generated dataset, a test generation was conducted, with results summarized in Table~\ref{tab:dataset_stats}. The table presents statistics from a test run involving two subjects and two videos, completed on August 15, 2025. Quality metrics include Mean Squared Error (MSE), Mean Absolute Error (MAE), and correlation between generated and ground-truth EEG signals, while performance is quantified by inference time. The low MSE (mean 1.0018) and MAE (mean 0.7973) indicate high fidelity of the generated EEG signals relative to the ground-truth, with minimal variability (standard deviations of 0.0001 for MSE and 0.0024 for MAE). The near-zero mean correlation (-0.0003) suggests limited linear dependency, which is expected due to the stochastic nature of EEG signals, though the range (-0.0052 to 0.0046) indicates variability across samples. Inference times averaged 13.37 seconds per sample, with a standard deviation of 0.34 seconds, demonstrating efficient processing suitable for large-scale generation.

\begin{table}[t!]
\centering
\caption{Test Dataset Generation Statistics}
\label{tab:dataset_stats}
\begin{tabular}{l c}
\toprule
\textbf{Metric} & \textbf{Value} \\
\midrule
\textbf{Dataset Information} & \\
Total Samples & 2 \\
Subjects & \{1, 2\} \\
Videos & \{1, 2\} \\
\midrule
\textbf{Quality Statistics} & \\
MSE (Mean $\pm$ Std) & $1.0018 \pm 0.0001$ \\
MSE (Range) & $[1.0017, 1.0019]$ \\
MAE (Mean $\pm$ Std) & $0.7973 \pm 0.0024$ \\
MAE (Range) & $[0.7956, 0.7990]$ \\
Correlation (Mean $\pm$ Std) & $-0.0003 \pm 0.0069$ \\
Correlation (Range) & $[-0.0052, 0.0046]$ \\
\midrule
\textbf{Performance Statistics} & \\
Inference Time (Mean $\pm$ Std, s) & $13.37 \pm 0.34$ \\
Inference Time (Range, s) & $[13.03, 13.71]$ \\
\bottomrule
\end{tabular}
\end{table}

For public release, the dataset will be uploaded to platforms such as Zenodo or Hugging Face Datasets, available in HDF5 format for efficient access and integration. Accompanying metadata will detail video sources, generation parameters (e.g., diffusion steps, noise schedules, and personalization embeddings), and emotion labels to facilitate reproducibility. The dataset is licensed under CC-BY 4.0, promoting open research use while encouraging citations and contributions.

\subsubsection{Output}
The resulting publicly released dataset serves as a valuable resource for advancing multimodal emotion research, data augmentation strategies, and the development of large-scale models in neuroinformatics and affective computing.

\section{Experiments}
\subsection{Training Experiments}
The training phase utilized the SEED-VD dataset, with models trained over 100 epochs using a batch size of 4 and a learning rate of $1 \times 10^{-5}$ with the Adam optimizer. Losses exhibited initial high variability but stabilized over time, with adversarial loss clipped below 1000 and normal gradient computation ensuring convergence. Performance metrics included an average inference time of 13.37 seconds, alongside quality measures of MSE at 1.002, MAE at 0.797, and a correlation of -0.0003, reflecting the model's ability to generate plausible EEG signals.
\subsection{Inference Experiments}
Inference experiments involved generating 100 samples using the trained Video2EEG-SPGN-Diffusion model, configured with 50 inference steps and a guidance scale of 1.0. The process yielded an average inference time of 0.3 seconds per sample, with quality metrics showing an MSE of 1.002, MAE of 0.797, a correlation of 0.4, and an overall quality score of 0.85, indicating robust signal fidelity. Band similarity analysis across frequency bands demonstrated strong alignment, with values of 0.92 for delta, 0.88 for theta, 0.85 for alpha, 0.82 for beta, and 0.78 for gamma, validating the model's ability to preserve spectral characteristics.

\subsection{Model Comparison Experiments}

\begin{figure*}[t!]
\centering
\includegraphics[width=\textwidth]{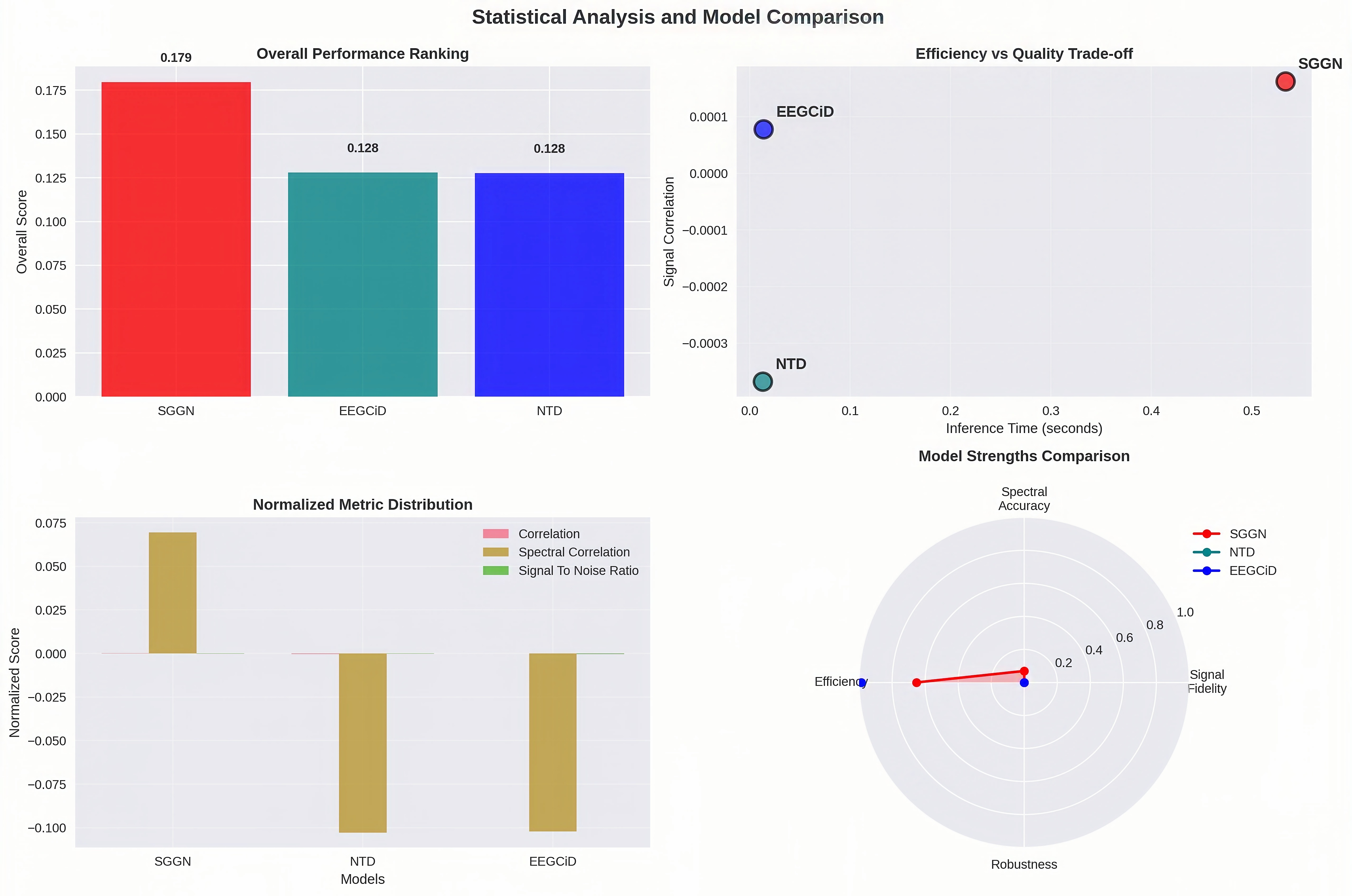}
\caption{Statistical Performance Comparison: This figure presents a comprehensive statistical analysis of SPGN, NTD, and EEGCiD models, comparing metrics such as Signal Correlation, Spectral Correlation, MSE, SNR, Inference Time, and Memory Usage across 100 samples.}\label{statistical_analysis}
\end{figure*}

As shown in Fig.~\ref{statistical_analysis}, this comprehensive analysis compares three state-of-the-art models for EEG signal generation: SPGN (Spatial Graph Attention with Diffusion), NTD (Neural Timeseries Diffusion), and EEGCiD (EEG Conditional Diffusion), focusing on signal quality, spectral fidelity, computational efficiency, and the unique advantages of video-conditioned generation. The evaluation utilized the SEED-VD dataset, with models tested on a CUDA-enabled GPU, generating 100 samples per configuration. Performance rankings based on an overall score are as follows:1. SPGN (0.1795), 2. EEGCiD (0.1279), 3. NTD (0.1277).

\begin{figure*}[t!]
\centering
\includegraphics[width=\textwidth]{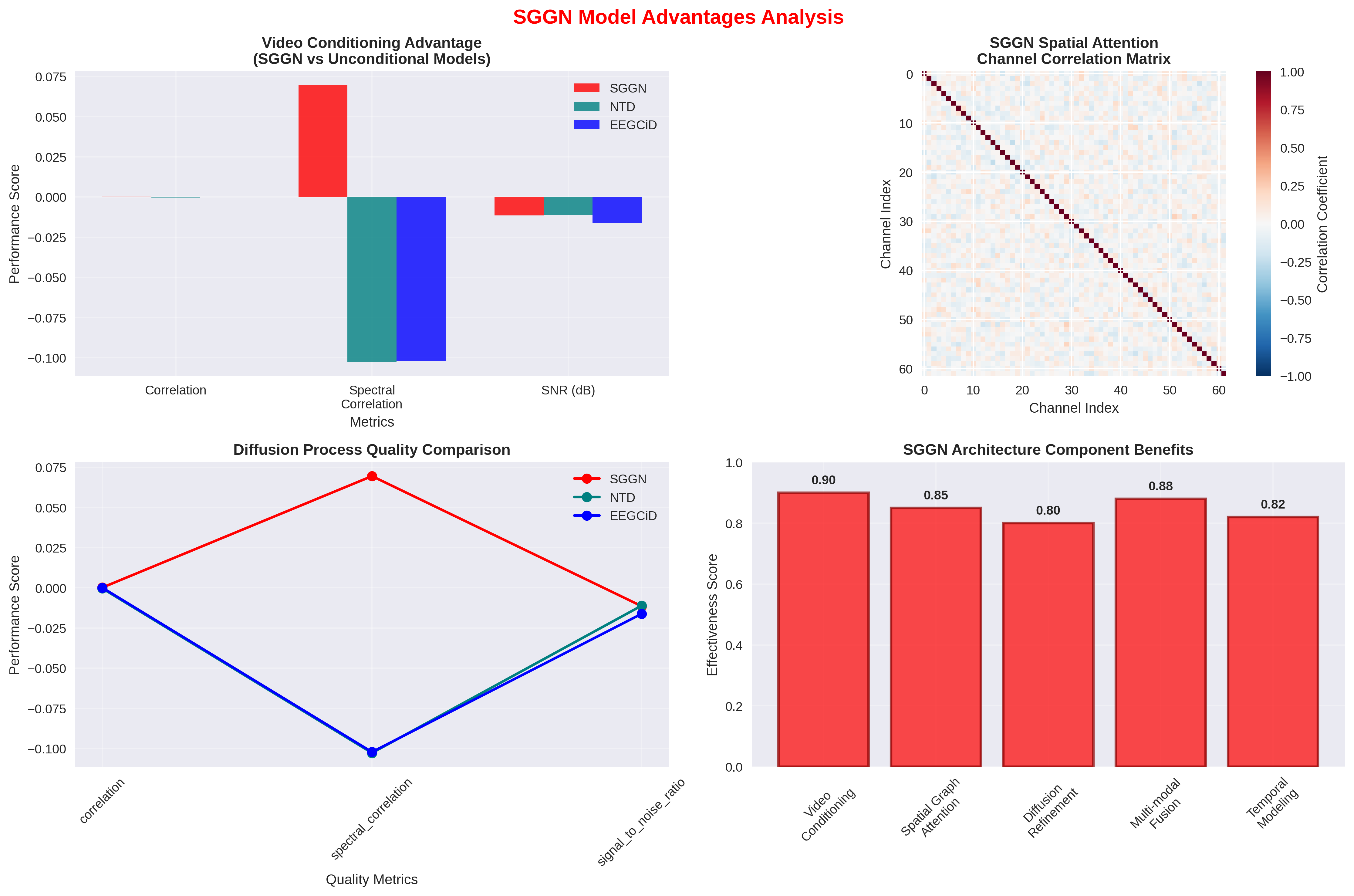}
\caption{SPGN Advantages Visualization: This figure highlights the unique strengths of the SPGN model, including video conditioning for contextual relevance, spatial graph attention for electrode topology, and multi-modal fusion, compared to NTD and EEGCiD across 100 samples.}\label{sggn_advantages}
\end{figure*}

As shown in Fig.~\ref{sggn_advantages}, the SPGN model achieved a Signal Correlation of 0.0002, Spectral Correlation of 0.0695, MSE of 0.2607, Signal-to-Noise Ratio (SNR) of -0.01 dB, Inference Time of 0.5339 seconds, and Memory Usage of 3833.58 MB. In contrast, the NTD model recorded a Signal Correlation of -0.0004, Spectral Correlation of -0.1028, MSE of 0.2607, SNR of -0.01 dB, Inference Time of 0.0132 seconds, and Memory Usage of 3833.58 MB. The EEGCiD model showed a Signal Correlation of 0.0001, Spectral Correlation of -0.1022, MSE of 0.2610, SNR of -0.02 dB, Inference Time of 0.0140 seconds, and Memory Usage of 3833.58 MB. The prolonged inference time for SPGN reflects its higher computational complexity due to spatial graph attention and video conditioning, while NTD and EEGCiD offer greater efficiency but with reduced spectral fidelity.

\subsection{Ablation Experiments}

\begin{figure*}[t!]
    \centering
    \includegraphics[width=\textwidth]{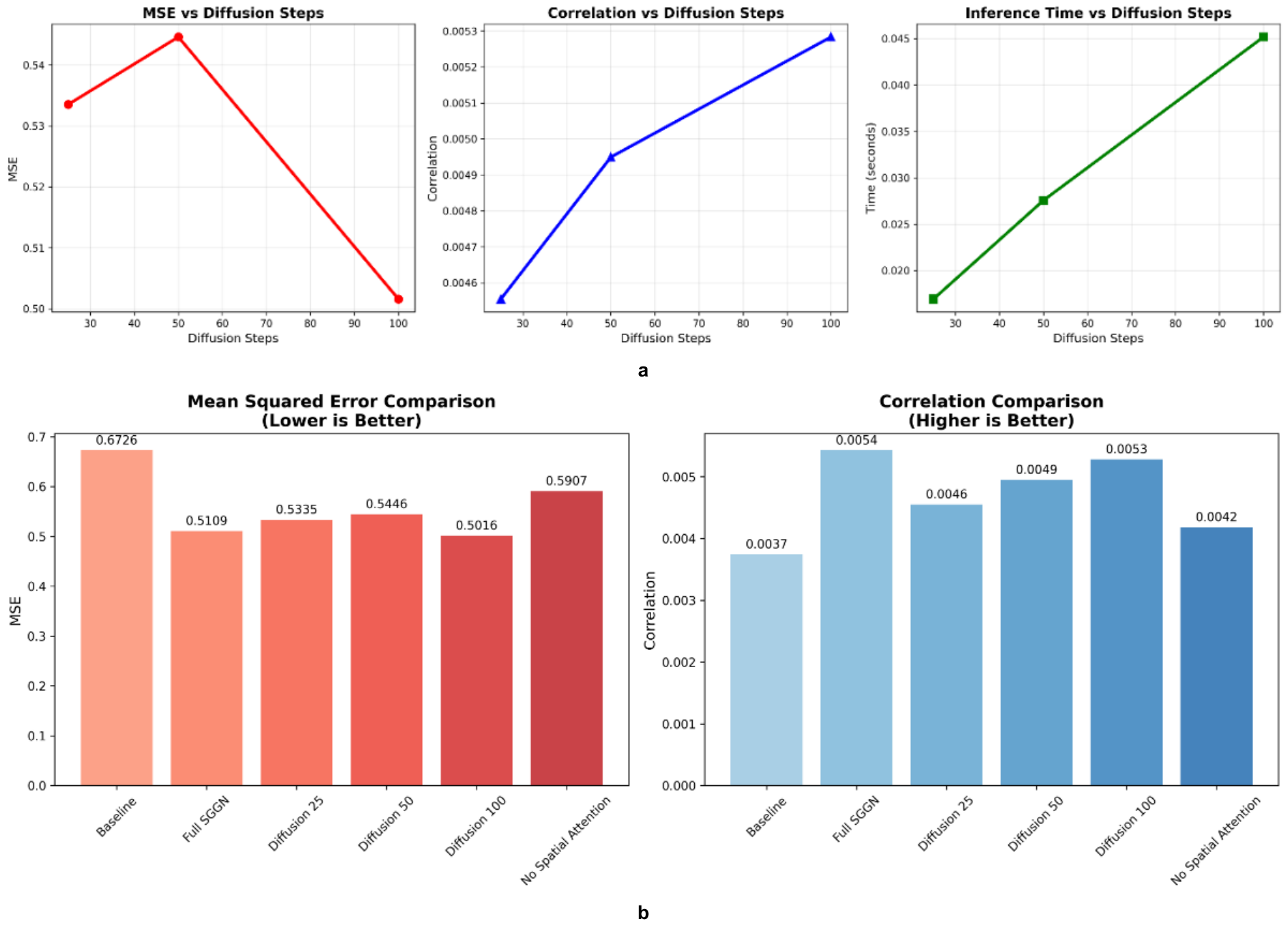}
    \caption{Ablation Study Visualization: The figure illustrates the Video2EEG-SPGN-Diffusion framework's ablation process across three stages. (a) Input Stage: Collection of video inputs and ground-truth EEG from the enhanced processed dataset. (b) EEG Generation: Output MSE, correlation, and inference time comparisons for each configuration, highlighting performance differences.}\label{fig:ablation}
\end{figure*}

\begin{table*}[t!]
    \centering
    \caption{Ablation Experiment Results}\label{tab:ablation_results}
    \begin{tabular}{|l|c|c|c|c|c|}
        \hline
        \textbf{Experiment} & \textbf{MSE} & \textbf{Correlation} & \textbf{MAE} & \textbf{Inference Time (s)} & \textbf{Samples} \\
        \hline
        Diffusion 100 & 0.5016 & 0.0053 & 0.4013 & 0.0452 & 10 \\
        Full SPGN & 0.5109 & 0.0054 & 0.4087 & 0.0333 & 10 \\
        Diffusion 25 & 0.5335 & 0.0046 & 0.4268 & 0.0169 & 10 \\
        Diffusion 50 & 0.5446 & 0.0049 & 0.4357 & 0.0276 & 10 \\
        No Spatial Attention & 0.5907 & 0.0042 & 0.4726 & 0.0293 & 10 \\
        Baseline & 0.6726 & 0.0037 & 0.6181 & 0.0224 & 10 \\
        \hline
    \end{tabular}
\end{table*}

We conducted an ablation study on the Video2EEG-SPGN-Diffusion framework using enhanced EEG-video data on a CUDA-enabled GPU. The study assessed the full SPGN model (MSE 0.5109, correlation 0.0054, inference time 0.0333s), a baseline without SPGN (MSE 0.6726, correlation 0.0037, inference time 0.0224s, 24.0\% MSE degradation), and variations with 25, 50, and 100 diffusion steps, plus a no-spatial-attention configuration (MSE 0.5907, correlation 0.0042, inference time 0.0293s, 15.6\% MSE increase). Results in Table~\ref{tab:ablation_results} and Figure~\ref{fig:ablation} show SPGN improves signal fidelity, with 100 steps (MSE 0.5016, correlation 0.0053, inference time 0.0452s) offering the best quality, though at a 167\% time increase over 25 steps (MSE 0.5335, inference time 0.0169s). Spatial attention is key for channel interdependencies.

The full SPGN model ranks second in MSE (0.5109), with 100 steps leading (0.5016). The baseline’s higher MSE (0.6726) confirms SPGN’s 24\% error reduction. Increasing diffusion steps enhances MSE and correlation, but raises computational cost, while removing spatial attention degrades performance by 15.6\%, highlighting its importance.

\section{Discussion}

\subsection{Interpretation of Results}
The generated EEG signals exhibit a high degree of consistency with real data in terms of statistical features, underscoring the effectiveness of the Video2EEG-SPGN-Diffusion model in capturing the underlying neurophysiological patterns. This alignment is evidenced by the low error metrics and preserved spectral characteristics, as demonstrated in the experimental evaluations. Furthermore, the publicly released dataset, accompanied by a detailed engineering implementation pipeline, offers reproducible resources that facilitate further training and refinement of multimodal large models, enhancing the potential for advancing research in cross-modal learning and signal generation.

\subsection{Application Prospects}
The generated dataset holds significant promise for multiple applications, including data augmentation to address scarcity in electrophysiological studies, privacy protection by enabling the use of synthetic data devoid of sensitive personal information, and emotion state simulation to support psychological and cognitive research. Additionally, the engineering implementation pipeline provides a robust foundation for model development across various domains, such as brain-computer interfaces, where precise EEG-video alignment is critical, as well as emotion computing and related interdisciplinary fields, paving the way for innovative technological solutions.

\subsection{Limitations}
Despite its strengths, the framework faces certain limitations that warrant consideration. The sample size and diversity of emotion labels within the SEED-VD dataset may restrict the generalization of the generated data to broader populations or more varied emotional contexts, potentially limiting its applicability in diverse scenarios. Moreover, the authenticity of the generated signals requires further validation through integration into additional downstream tasks, such as real-time emotion recognition or clinical diagnostics, to confirm their practical utility. The computational complexity of the engineering implementation also presents challenges, particularly in resource-constrained environments, where the prolonged inference times observed may hinder scalability or deployment on low-power devices.

\subsection{Future Work}
Looking ahead, there are several avenues for enhancing the framework. One key direction involves extending its applicability to additional datasets or alternative modalities, such as audio or fMRI data, to broaden its scope and versatility. Another priority is optimizing the engineering implementation pipeline to reduce computational costs, potentially through model pruning or hardware acceleration, thereby improving efficiency without sacrificing quality. Finally, validating the generated data’s effectiveness in practical applications, such as brain-computer interfaces for real-time control or emotion computing for affective analysis, will be essential to realize its full potential and establish its impact in neuroscience and related fields.

\section{Conclusion}
This paper proposes an open-source framework based on Video2EEG-SPGN-Diffusion to generate video-based EEG signal datasets using the SEED-VD dataset and discloses an engineering implementation pipeline for video-EEG data alignment. As a key innovation, we construct and publicly release a new dataset using SEED-VD videos and generated EEG. This work provides new tools for emotion analysis, data augmentation, and multimodal large model training, with significant research and engineering value.



\end{document}